\documentclass{article}

\PassOptionsToPackage{numbers, compress}{natbib}
\usepackage[main,final]{neurips_2026}

\usepackage[utf8]{inputenc} 
\usepackage[T1]{fontenc}    
\usepackage{hyperref}       
\usepackage{url}            
\usepackage{booktabs}       
\usepackage{amsfonts}       
\usepackage{amsmath}        
\usepackage{nicefrac}       
\usepackage{microtype}      
\usepackage{xcolor}         
\usepackage{graphicx}       
\usepackage{wrapfig}        
\usepackage{multirow}       
\usepackage{makecell}       
\usepackage{placeins}
\usepackage{needspace}
\usepackage{enumitem}
\usepackage{titletoc}
\title{EventLens: Event-Structure Reinforcement Learning for Video Understanding}

\author{%
  \textbf{Guowen Zhang}$^{1}$ \quad \textbf{Boshen Xu}$^{1}$ \quad
  \textbf{Zihao Yue}$^{1}$ \quad \textbf{Ziheng Wang}$^{1}$ \\
\textbf{Xiaokun Liu}$^{2*}$ \quad \textbf{Xin Tao}$^{2}$ \quad
  \textbf{Wenyu Qin}$^{2}$ \quad \textbf{Pengfei Wan}$^{2}$ \quad
  \textbf{Qin Jin}$^{1\dagger}$ \\
  \\
  {\normalfont $^{1}$AIM3 Lab, Renmin University of China \quad $^{2}$Kling AI}
\\ 
    Project Page: \url{https://lostgreen.github.io/EventLens/}
}

\begin{document}

\renewcommand{\thefootnote}%
{\fnsymbol{footnote}}
\footnotetext[0]{$\dagger$ Corresponding Author; $^*$ Industry Project Lead.}

\maketitle

\begin{abstract}
Recent multimodal large language models (MLLMs) achieve strong results on video question answering, yet often fail to recover the temporal structure that makes a video coherent. They may merge neighboring events, miss single-event progression, or infer implausible relations across events. We study structured temporal understanding as the ability to recover event structure from visual evidence, including event boundaries, single-event progression, and multi-event relations. We argue that this gap arises from a mismatch between current post-training objectives and temporal reasoning, where caption-style supervision allows models to rely on language shortcuts rather than visual temporal grounding.
To address this, we propose \textbf{EventLens}, a vision-centric reinforcement learning framework that learns video understanding through \emph{event-structure recovery}.
Instead of reconstructing captions, EventLens trains models to recover temporal structure under controlled perturbations, such as altered temporal granularity, reversed progression, and shuffled event order.
We instantiate this framework with a three-level temporal ontology and derive three verifiable task families: event segmentation, progression discrimination, and multi-event relation reasoning.
These tasks admit deterministic rewards, enabling task-specialized GRPO training without human preference labels or learned reward models.
We further introduce multi-teacher on-policy distillation (MT-OPD) to consolidate specialized policies into a unified model.
Experiments on Qwen3-VL backbones show consistent improvements on temporal grounding and event-sensitive reasoning benchmarks, while preserving general video-QA performance.
Learning curves further demonstrate that EventLens achieves comparable downstream transfer with significantly reduced training cost compared to direct mixed RL optimization.
\end{abstract}

\section{Introduction}
\label{sec:intro}
Modern multimodal large language models (MLLMs) have achieved strong performance on a wide range of video question answering benchmarks~\cite{mlvu2025,longvideobench2024,videomme2025}.
However, this success largely reflects the ability to recognize what is visible, rather than to understand how events unfold over time.
When tasks require explicit temporal structure, models often fail to separate events, track their progression, or reason about relations across them.
As a result, models that appear capable on coarse-grained benchmarks can break down when deeper temporal understanding is required.

We study this limitation as \emph{structured temporal understanding}: the ability to recover event boundaries, single-event progression, and multi-event relations from visual evidence.
This capability underlies tasks such as temporal localization, progression analysis, and multi-event reasoning~\cite{mvbench2024,mvpbench2026}.
Unlike frame-level recognition or caption matching, these tasks require recovering the underlying event structure of a video.

Recent evidence suggests that current models often succeed without such structure.
Performance drops sharply when evaluation moves from single-event to multi-event settings~\cite{vector2025}, and models frequently perform near chance on event reordering tasks where humans achieve near-perfect accuracy~\cite{mvpbench2026}.
More fundamentally, models can answer a substantial portion of video QA questions without accessing visual input~\cite{breakingdown2025}, and show limited sensitivity to temporal perturbations such as frame shuffling~\cite{tvbench2025,tempcore2025}.
Timestamp prediction models often output similar time ranges across unrelated videos, indicating reliance on language patterns rather than visual grounding~\cite{videoexpert2025}.
These observations indicate that current evaluation protocols often conflate temporal understanding with language priors and world knowledge.

We argue that this gap stems from a mismatch between current post-training objectives and structured temporal understanding.
Most approaches rely on text-supervised objectives, where models are trained to generate captions or answers aligned with language distributions.
Such objectives allow models to exploit language shortcuts rather than grounding predictions in visual temporal evidence~\cite{breakingdown2025,tvbench2025}.
Meanwhile, existing temporal learning signals remain fragmented across isolated skills.
Prior work has explored temporal grounding~\cite{timer1_2025,timelens2025,tempr1_2025,museg2025}, event prediction~\cite{tempura2025,nep2025}, and temporal reasoning for QA~\cite{videor1_2025}, while classical self-supervised tasks such as VCP~\cite{vcp2020}, OPN~\cite{opn2017}, and temporal transformation recognition~\cite{jenni2020temporal} were designed for representation learning rather than post-training MLLMs.
However, no existing objective explicitly trains models to jointly recover event boundaries, intra-event progression, and inter-event relations as a unified capability.

\begin{figure*}[t]
  \centering
  \includegraphics[width=\textwidth]{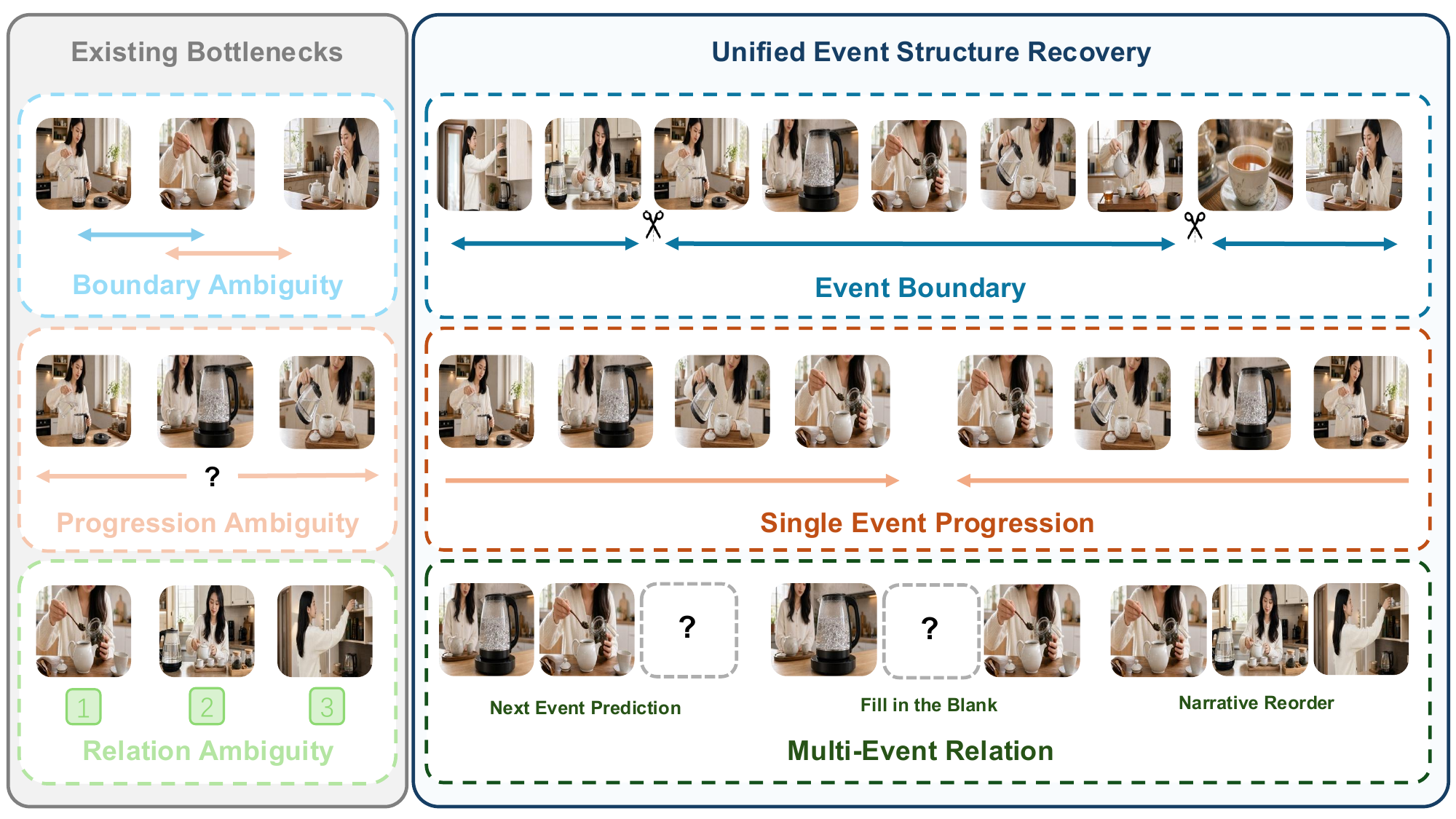}
  \caption{Motivation and task view of unified event-structure recovery.
  Existing video MLLMs often suffer from boundary ambiguity, progression ambiguity,
  and relation ambiguity when they must recover explicit temporal structure.
  We address these bottlenecks with three event-structure recovery tasks:
  event-boundary recovery, single-event progression discrimination, and
  multi-event relation reasoning through next-event prediction, fill-in, and narrative reordering.}
  \label{fig:teaser}
\end{figure*}

To address this limitation, we propose \emph{EventLens}, a vision-centric reinforcement learning framework that learns video understanding through \emph{event-structure recovery}.
The key idea is to train models not to match language descriptions, but to recover temporal structure from visual evidence.
We operationalize this by deliberately perturbing temporal structure through changes in granularity, playback direction, and event order, and training the model to recover the original structure.
These recovery tasks are inherently vision-centric and admit deterministic verification signals, enabling reinforcement learning without additional human preference labels or learned reward models.

To systematically instantiate this framework, we introduce a three-level temporal ontology consisting of thematic phases, semantic events, and atomic action units.
Based on this ontology, we derive three complementary task families:
event segmentation, progression discrimination, and multi-event relation reasoning via next-event prediction, fill-in, and clip reordering.
We further design a two-stage training paradigm that separates task specialization from policy unification.
We first train task-specific policies with GRPO under verifiable rewards, and then consolidate them into a unified model using multi-teacher on-policy distillation (MT-OPD), avoiding instability from directly mixing heterogeneous reward signals.

We evaluate EventLens on Qwen3-VL and Qwen2.5-VL backbones and compare it with supervised and reinforcement-learning post-training methods.
Across both Qwen-VL generations, EventLens improves temporal grounding and event-sensitive reasoning while maintaining competitive general video understanding.
Direct recovery-task evaluations show that EventLens integrates complementary segmentation, progression, and relation skills into a unified model.
Learning curve analysis further demonstrates that EventLens achieves comparable downstream transfer with substantially reduced training time compared to direct mixed RL optimization.

In summary, our contributions are as follows:
\begin{itemize}[leftmargin=*, itemsep=1pt, topsep=1pt, parsep=0pt]
\item We formulate structured temporal understanding as \emph{event-structure recovery} from visual evidence, unifying event boundaries, intra-event progression, and inter-event relations.

\item We propose \textbf{EventLens}, a vision-centric reinforcement learning framework that learns video understanding through temporal perturbation and recovery with verifiable rewards.

\item We introduce a training paradigm with task-specialized GRPO and multi-teacher on-policy distillation (MT-OPD), enabling stable integration of heterogeneous temporal skills.

\item We demonstrate consistent improvements on temporal-structure benchmarks with controlled retention on general video QA, and improved training efficiency compared to direct mixed RL optimization.
\end{itemize}

\section{Related work}
\label{sec:related}

\textbf{Video MLLMs and RL post-training.}
Recent video MLLMs, including LLaVA-Video~\cite{llavavideo2024}, VideoLLaMA3~\cite{videollama3_2025},
InternVL-2.5~\cite{internvl25_2025}, InternVideo2.5~\cite{internvideo25_2025},
Qwen2.5-VL~\cite{qwen25vl2025}, and Qwen3-VL~\cite{qwen3vl2025}, show strong
coarse-grained video understanding, but standard text-supervised post-training does not
directly optimize structured temporal recovery.
Recent RL methods introduce verifiable rewards for video QA, temporal perception, and
grounding~\cite{videor1_2025,arrowrl2025,timer1_2025,timelens2025,tempr1_2025,museg2025}.
VideoAuto-R1~\cite{videoautor1_2026} adapts reasoning to each query, while
OneThinker~\cite{onethinker2025} unifies image and video tasks through multi-task RL.
R1-Track~\cite{r1track2025} applies MLLM RL to object tracking;
ViewSAM~\cite{viewsam2026} learns cross-view referring tracking under weak supervision.
Closest to our use of proxy tasks, Visual Jigsaw~\cite{visualjigsaw2025},
VideoSSR~\cite{videossr2026}, ViSS-R1~\cite{vissr1_2025}, and
VidBridge-R1~\cite{vidbridger1_2026} train MLLMs with visual or temporal transformations.
These proxy-based methods establish the value of temporal perturbations for video MLLMs, but they
usually optimize one temporal skill, one perturbation family, or one downstream format.
Our framework instead uses a shared temporal abstraction to generate complementary recovery
tasks for event boundaries, single-event progression, and multi-event relations.

\textbf{Temporal structure learning and evaluation.}
Supervised temporal action segmentation and generic event boundary detection partition videos
into meaningful units, but they usually assume dense labels and one fixed
granularity~\cite{mstcn2019,ishikawa2021asrf,asformer2021,lu2024fact,gebd2021,tang2022ddm,jung2025onlinegebd}.
Event-centric benchmarks further test ordering, clue integration, causal logic, and temporal
semantic sensitivity~\cite{nextqa,eventbench2024,etbench2024,videoholmes2025,vector2025,rtime2024}.
Video-grounded entailment tree reasoning~\cite{vget2025} connects commonsense QA
to explicit verification of visual evidence.
Existing training objectives cover parts of this space through text-supervised event learning,
event prediction, or classical temporal pretext learning~\cite{tempura2025,timesuite2025,nep2025,aot_pickup2014,aot_wei2018,vcp2020,opn2017}.
We adapt these intuitions to MLLM RL post-training by making segmentation prompt-conditioned
across temporal levels and coupling it with single-event progression and multi-event relation reasoning under
deterministic verifiers.

\textbf{Distillation for multi-task post-training.}
Knowledge distillation and multiple-teacher distillation transfer complementary behaviors from
specialized teachers to one student~\cite{hinton2015distillingknowledgeneuralnetwork,wu2021teacherenoughpretrainedlanguage}.
On-policy distillation reduces train-inference mismatch by training on student-generated
trajectories~\cite{agarwal2024onpolicydistillationlanguagemodels}, and
Video-OPD~\cite{videoopd2026} applies this principle to temporal video grounding.
We use this family of techniques only as the consolidation mechanism: the main distinction is
that the teachers specialize in heterogeneous temporal recovery tasks whose rewards differ in
scale, sparsity, and response format.

\section{Method}
\label{sec:method}

\begin{figure*}[t]
  \centering
  \includegraphics[width=0.95\textwidth]{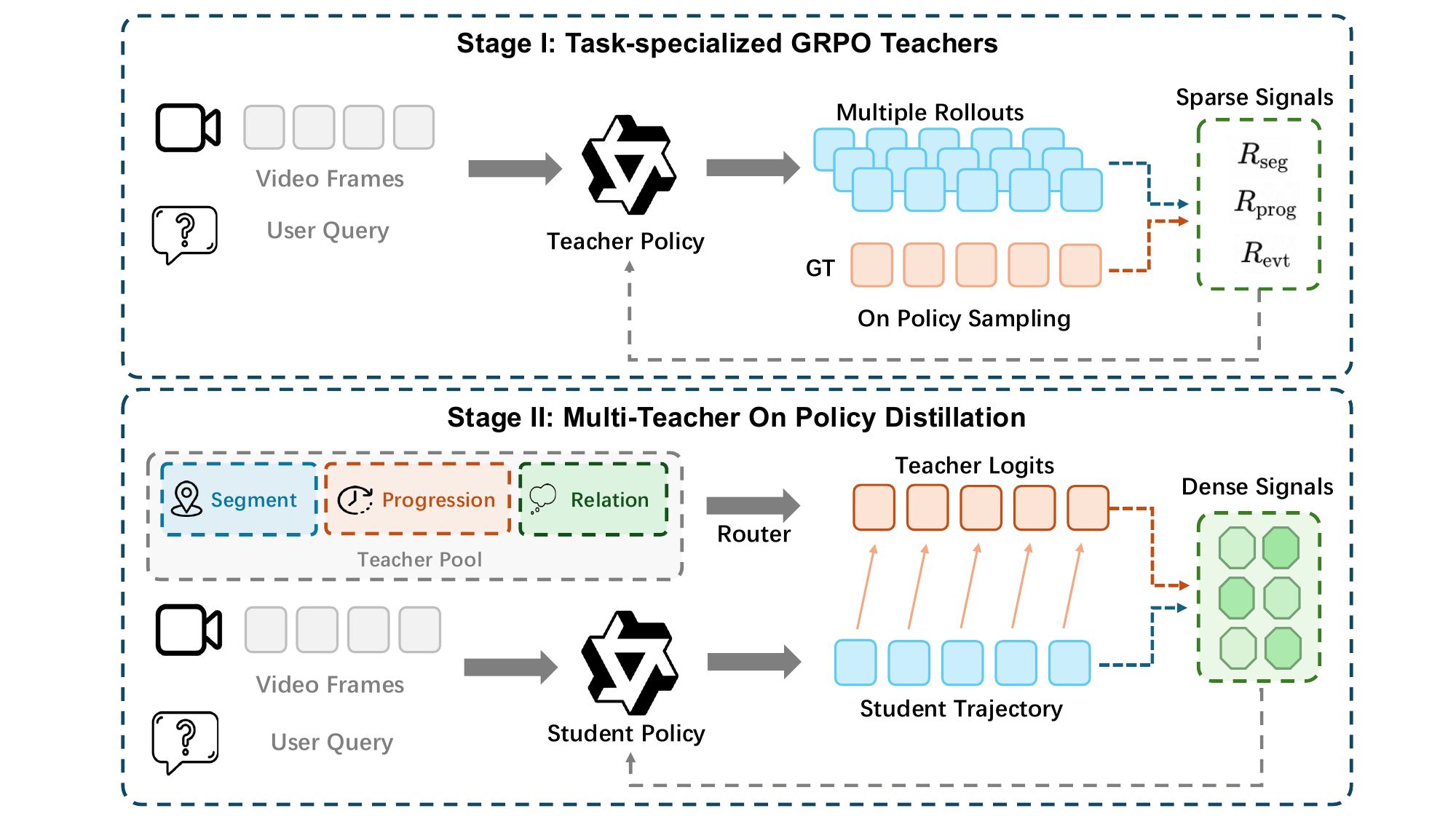}
  \caption{ Overview of EventLens. EventLens adopts a two-stage training pipeline for event-structure
reinforcement learning. In Stage I, deterministic rewards train task-specialized GRPO teachers
for event-boundary localization, single-event progression discrimination, and multi-event relation
reasoning. In Stage II, MT-OPD routes each student trajectory to the corresponding teacher and
distills dense token-level teacher logits into one deployable student. The teacher is selected by
recovery family, so each student trajectory is supervised by the teacher trained with the matching
verifier.}
  \label{fig:eventlens_pipeline}
\end{figure*}


EventLens formulates structured video understanding as recovering latent temporal structure from visual evidence.
For a source video, the model must recover three complementary components of event structure:
event boundaries, single-event progression, and multi-event relations.
The framework constructs three event-structure recovery task families from two publicly available
video corpora:
\textbf{(i) level-conditioned temporal segmentation},
\textbf{(ii) single-event progression discrimination}, and
\textbf{(iii) multi-event relation reasoning}.
All three families are derived from a shared temporal ontology.
Figure~\ref{fig:eventlens_pipeline} summarizes how these instances feed the two-stage
training pipeline.
EventLens first uses these deterministic verifiers to train task-specialized GRPO teachers and then merges
their skills into one deployable student with multi-teacher on-policy distillation.
We denote the corresponding reward families by
$\mathcal{R}=\{R_{\text{seg}},R_{\text{prog}},R_{\text{evt}}\}$, where $R_{\text{seg}}$ corresponds to temporal segmentation,
$R_{\text{prog}}$ to single-event progression discrimination,
and $R_{\text{evt}}$ to multi-event relation reasoning.
Each reward is computed directly from the constructed target structure,
so training does not require additional human preference labels or learned reward models.

\subsection{Hierarchical temporal ontology and corpus construction}
\label{sec:boundary}

\paragraph{Three-level temporal ontology.}
We organize each video with a three-level temporal ontology:
L1 (thematic phase) denotes a maximal contiguous interval whose constituent events
share a common high-level intent, topic, or discourse function;
L2 (semantic event) denotes a coherent activity under relatively stable participants,
location, and short-term goal, potentially spanning multiple shots if they preserve spatio-temporal continuity;
and L3 (atomic action unit) denotes the shortest interval that still expresses
a visually identifiable motion or state transition.
The hierarchy moves from coarse discourse structure to fine-grained motion.
All recovery tasks share this ontology, so they agree on what counts as a meaningful temporal unit.

\Needspace{0.5\textheight}
\paragraph{Corpus construction.}
We select source videos from TimeLens-100K~\cite{timelens2025} and ET-Instruct-164K~\cite{etbench2024}.
For both sources, we use only their official training partitions and exclude videos and
annotations from the held-out test splits used by our evaluation benchmarks.
From this parent pool, we filter videos by duration and quality.
We sample frames at 1 FPS and use TransNetV2 to obtain candidate shot intervals.

\par\smallskip
\noindent\begin{minipage}{\linewidth}
  \centering
  \includegraphics[width=\linewidth]{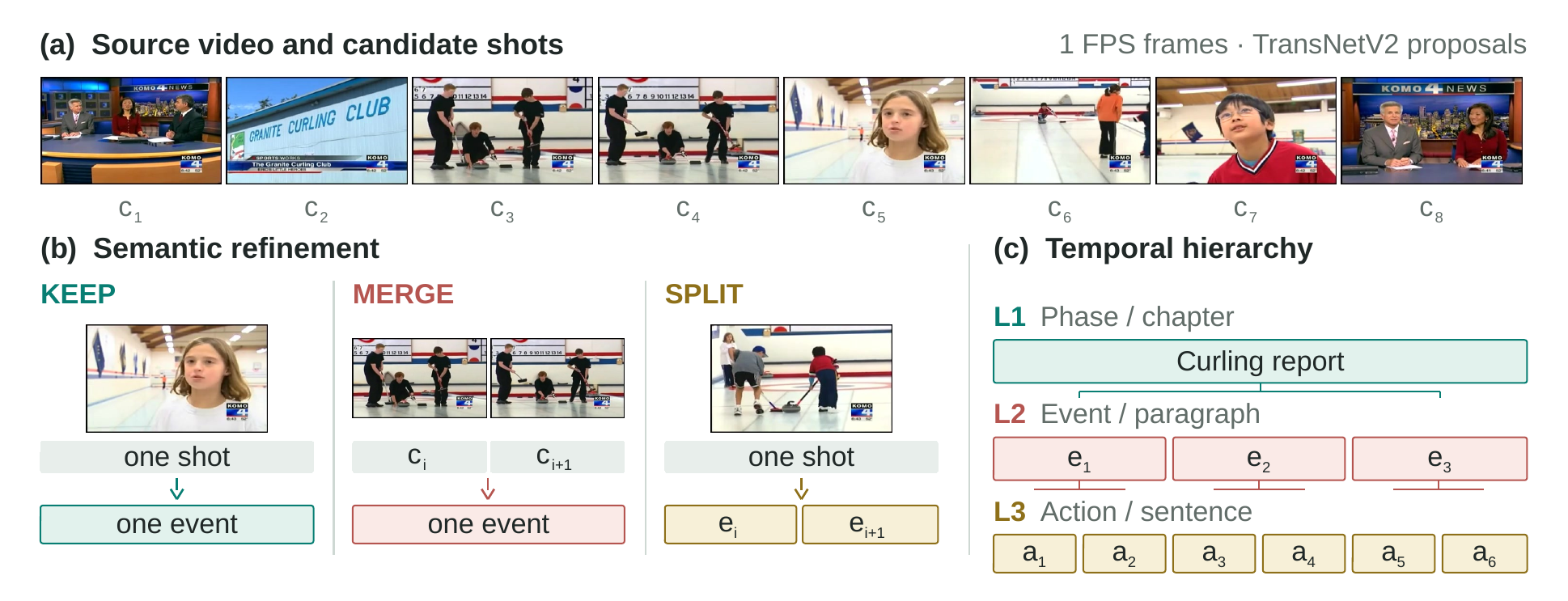}
  \makeatletter\def\@captype{figure}\makeatother
  \caption{Scene-first construction of event structure for a curling news report.
  (a) Sampled frames and candidate shots. (b) KEEP retains a shot as one event,
  MERGE combines consecutive shots into one event, and SPLIT introduces event
  boundaries within a shot. (c) L1 phases contain L2 events, which contain L3
  actions. Frames come from the historical report illustration; boundaries and
  event assignments are schematic.}
  \label{fig:construction_main}
  \label{fig:cr_construction}
\end{minipage}\par\smallskip

\paragraph{Semantic refinement and hierarchy.}
The shot proposals are refined according to semantic continuity: consecutive shots depicting a continuous event are merged, while a shot containing multiple semantic transitions is split into distinct events (Fig.~\ref{fig:construction_main}).
The resulting semantic events define L2 intervals, each containing a sequence of nested L3 action units that capture atomic motion or state transitions.
Adjacent L2 events sharing a high-level intent, topic, or discourse function are grouped into L1 phases.
We enforce scene coverage, temporal validity, and parent--child containment before converting the hierarchy into recovery-task targets.
Annotation details and representative examples are provided in Appendix~\ref{app:cr_data_quality}.

\paragraph{Recovery instance construction.}
From the resulting hierarchy, we obtain an ordered event chain $E=(e_1,\dots,e_M)$
at a chosen ontology level.
The same filtered video can yield segmentation targets at different semantic levels,
event-level clips for progression discrimination, and multi-event windows for temporal-dependency reasoning.
Granularity is specified through the prompt rather than the architecture,
so the same MLLM can be trained to reason over phases, events, or action units within one framework.
Each instance is associated with one recovery family, which determines both the verifier and
the teacher used during multi-teacher OPD.

\subsection{Event-structure recovery tasks}
\label{sec:proxy}

\paragraph{Hierarchical temporal segmentation.}
This task trains the model to recover event boundaries at the semantic scale specified by the
prompt.
Given a video and a granularity-specifying prompt, the model outputs temporal segment boundaries
at the requested level.
Each training instance requests exactly one level, forcing the model to learn what constitutes
a meaningful unit at that semantic scale rather than localizing a fixed text query.
Because segmentation is set-valued and the number of predicted segments may differ from the target,
we first compute an optimal one-to-one matching between predicted and ground-truth segments.
Let $\mathcal{S}^{*}=\{s^{*}_j\}_{j=1}^{N^{*}}$ and
$\hat{\mathcal{S}}=\{\hat{s}_i\}_{i=1}^{N}$ denote the ground-truth and predicted segment sets.
For two temporal intervals $a$ and $b$, let
$\operatorname{tIoU}(a,b)=|a\cap b|/|a\cup b|$, where $|\cdot|$ denotes duration.
We obtain the matched pairs by
\begin{equation}
  \mathcal{M}^{*}
  =
  \arg\max_{\mathcal{M}\in\Omega}
  \sum_{(i,j)\in\mathcal{M}}
  \operatorname{tIoU}(\hat{s}_i, s^{*}_j)
\end{equation}
where $\Omega$ is the set of all one-to-one partial matchings between
$\hat{\mathcal{S}}$ and $\mathcal{S}^{*}$.
We then score the prediction with an F1-style reward that penalizes both over- and under-segmentation:
\begin{equation}
  \begin{gathered}
  R_{\text{seg}}
  =
  \frac{2 P_{\text{seg}} C_{\text{seg}}}
       {P_{\text{seg}} + C_{\text{seg}} + \epsilon}, \\
  C_{\text{seg}}
  =
  \frac{1}{N^{*}}
  \sum_{(i,j)\in\mathcal{M}^{*}}
  \operatorname{tIoU}(\hat{s}_i, s^{*}_j),
  \quad
  P_{\text{seg}}
  =
  \frac{1}{N}
  \sum_{(i,j)\in\mathcal{M}^{*}}
  \operatorname{tIoU}(\hat{s}_i, s^{*}_j).
  \end{gathered}
  \label{eq:seg_reward}
\end{equation}
where $C_{\text{seg}}$ measures ground-truth coverage, $P_{\text{seg}}$ measures prediction
precision, and $\epsilon$ is a small constant for numerical stability.
If no valid segment can be parsed from the model response, we set $R_{\text{seg}}=0$.

\paragraph{Single-event progression discrimination.}
This task trains the model to recover the internal direction of a single event.
Given an event clip, we present either the original clip or its temporally reversed counterpart
and ask the model to predict the progression direction.
Reversed clips are generated on the fly and require no additional annotation.
Unlike classical temporal-direction pretexts on sub-second snippets~\cite{aot_wei2018},
we sample event-level clips (typically 5--30~s), so success requires recovering coherent
physical dynamics rather than relying on low-level motion artifacts.
The reward is binary correctness:
\begin{equation}
  R_{\text{prog}} = \mathbf{1}[\hat{y} = y^{*}]
\end{equation}
Here $y^{*}\in\{\mathrm{forward},\mathrm{reverse}\}$, and invalid predictions receive zero reward.

\paragraph{Multi-event relation reasoning.}
This task trains the model to recover dependencies among neighboring events.
To model temporal and semantic dependencies across events, let
$W=(e_1,\dots,e_m)$ denote a chronologically ordered window of events.
We construct three complementary tasks from $W$.
In next-event prediction, the model observes the visual prefix
$(e_1,\dots,e_i)$ and selects which candidate event should follow.
This tests whether the model can anticipate a plausible continuation from the preceding
visual process.
In fill-in, the model observes visual clips before and after a missing event and selects
the event that best connects the two sides; this requires reasoning about both preconditions
and consequences rather than only extrapolating forward.
In clip reordering, the same event clips are shuffled and the model outputs their
chronological permutation.
For the two selection tasks, candidates contain one correct answer and hard distractors drawn
from the same video or related event context, so the task cannot be solved by coarse topic
matching alone.
We present candidate events as visual clips with neutral identifiers rather than textual event
descriptions, which keeps the verifier tied to visual temporal evidence.
We treat these three tasks as one multi-event relation family with reward
\begin{equation}
  R_{\text{evt}} =
  \begin{cases}
    \mathbf{1}[\hat{y} = y^{*}], & \text{for next-event prediction and fill-in}, \\[0.3em]
    1 - \frac{\sum_{k=1}^{K}
    \left|\mathrm{pos}(k,\hat{\pi})-\mathrm{pos}(k,\pi^{*})\right|}
    {\sum_{k=1}^{K}\left|k-(K{+}1{-}k)\right|}, & \text{for clip reordering},
  \end{cases}
  \label{eq:event_reward}
\end{equation}
where $\hat{\pi}$ and $\pi^{*}$ denote the predicted and ground-truth permutations,
and $\mathrm{pos}(k,\pi)$ is the position of clip $k$ in permutation $\pi$ over $K\ge2$ clips.
The first case gives exact supervision for discrete event selection.
The second case uses a normalized temporal-jigsaw displacement reward,
following the same motivation behind smooth order-aware supervision in prior video RL work
such as VideoSSR~\cite{videossr2026},
so partially correct orderings still provide dense learning signal.
Malformed selections or permutations receive zero reward.

\subsection{Task-specialized GRPO teachers}
\label{sec:grpo}

All event-structure recovery tasks emit deterministic rewards computed directly from task construction,
making GRPO~\cite{deepseekr1} a suitable optimizer for task-specialized teacher training.
We train one teacher for each reward family
$t \in \mathcal{T}=\{\mathrm{seg}, \mathrm{prog}, \mathrm{evt}\}$.
For an input $x$ from task family $t$, the teacher samples a group of responses,
the verifier $R_t$ scores each response, and GRPO updates the policy with the standard
group-normalized, KL-regularized objective.
We summarize the teacher update with one compact objective:
\begin{equation}
  \mathcal{J}^{(t)}_{\mathrm{GRPO}}
  =
  \mathbb{E}\!\left[
    \frac{1}{G}\sum_{i=1}^{G}\rho_i^{(t)}(\theta_t)\hat{A}_i^{(t)}
    - \beta D_{\mathrm{KL}}(\pi_{\theta_t}\|\pi_{\mathrm{ref}})
  \right]
  \label{eq:grpo_teacher}
\end{equation}
where the expectation is over $x\sim\mathcal{D}_t$ and response groups
$o_i\sim\pi_{\theta_{t,\mathrm{old}}}(\cdot\mid x)$;
$\rho_i^{(t)}(\theta_t)=\pi_{\theta_t}(o_i\mid x)/\pi_{\theta_{t,\mathrm{old}}}(o_i\mid x)$,
and $\hat{A}_i^{(t)}$ is the group-normalized advantage computed from verifier scores
$R_t(o_i,x)$.
For task-composition ablations, we additionally train mixed-GRPO variants by sampling multiple
reward families in one run.
Because segmentation, single-event progression, and multi-event relation rewards have different scales and
variance profiles, these mixed-GRPO baselines use EMA-GRPO~\cite{onethinker2025}, which maintains
task-wise exponential moving averages of reward statistics for advantage normalization.

\subsection{EventLens teacher consolidation}
\label{sec:opd}

Directly optimizing a mixed GRPO objective couples all reward families inside the same
high-variance policy-gradient update.
EventLens instead distills the task-specialized teachers into a single student
with on-policy distillation~\cite{agarwal2024onpolicydistillationlanguagemodels},
building on classical knowledge distillation~\cite{hinton2015distillingknowledgeneuralnetwork}
and multiple-teacher distillation~\cite{wu2021teacherenoughpretrainedlanguage}.
For an input $x$ sampled from task-family pool $\mathcal{D}_t$, the current student first
generates an on-policy response.
We then route the same trajectory to the corresponding teacher:
segmentation examples use the segmentation teacher, single-event progression examples use the progression teacher,
and next-event, fill-in, and reordering examples use the multi-event relation teacher.
The routed teacher provides top-$k$ next-token distributions on the student trajectory, and the
student minimizes a sparse forward KL on this support:
\begin{equation}
  \mathcal{L}_{\mathrm{MT\text{-}OPD}}
  =
  \mathbb{E}_{t\sim q,\;x\sim\mathcal{D}_t,\;y\sim\pi_\theta(\cdot\mid x)}
  \left[\widehat{D}^{(k)}_{\mathrm{KL}}(\pi_{T_t}\|\pi_\theta; x,y)\right],
  \label{eq:mt_opd_compact}
\end{equation}
where $q_t$ is the task-family sampling probability and
$\widehat{D}^{(k)}_{\mathrm{KL}}$ denotes the top-$k$ teacher-student KL along the student trajectory.
Both teacher and student probabilities are renormalized over the teacher top-$k$ set before the
KL is computed.
This trains the student on its own trajectories, reducing the train-inference mismatch of
offline teacher-output distillation while preserving each teacher's specialization in one
deployable model.

\section{Experiments}
\label{sec:exp}

\subsection{Experimental setup}
\label{sec:setup}

\paragraph{Implementation details.}
We build on Qwen3-VL-4B and Qwen3-VL-8B~\cite{qwen3vl2025}.
All post-training uses the proxy-task data described in Sec.~\ref{sec:boundary}.
Task-specialized teachers are trained with EMA-GRPO using learning rate
$5 \times 10^{-7}$, global batch size 64, rollout number $G{=}8$, and KL
coefficient $1 \times 10^{-2}$.
The final EventLens student is trained with multi-teacher OPD on the same mixed proxy-task pool,
with one student rollout per prompt, top-$k{=}10$ teacher distributions, and OPD KL
coefficient 1.0.
Training and evaluation use the same video budget: 256 frames and $256{\times}256$
maximum pixels.
For each benchmark, inference settings are matched across the pretrained and post-trained Qwen3-VL variants, using greedy decoding.
Temporal-grounding frame sampling is specified separately in Appendix~\ref{app:temporal_grounding_eval}.

\paragraph{Benchmarks and baselines.}
We organize evaluation by role rather than by dataset taxonomy.
Matched temporal grounding tests whether EventLens transfers to explicit boundary localization;
structured reasoning tests whether the same training helps event-level inference; and
general and long-video QA serve as retention checks.
Dataset names, metrics, and grouped results are reported in
Tables~\ref{tab:temporal} and~\ref{tab:general}.
For each Qwen3-VL backbone, we compare the pretrained checkpoint,
two text-supervised post-training variants (SFT-Desc and SFT-DVC), and EventLens
under the same inference protocol.
Appendix~\ref{app:temporal_grounding_eval} gives the temporal grounding prompt,
timestamp parser, and metric implementation, and Appendix~\ref{app:videossr_peer}
provides a same-scale reference point against a recent video post-training method.
We therefore interpret Tables~\ref{tab:temporal} and~\ref{tab:general} asymmetrically:
temporal grounding and event-sensitive reasoning are target evaluations, whereas broad
and long-video QA benchmarks are retention diagnostics rather than the primary
optimization target.

\subsection{Main results}
\label{sec:main_results}

\begin{table*}[t]
  \caption{Temporal grounding and complex reasoning under the shared 256-frame protocol.
  Temporal grounding uses mIoU; reasoning columns use official benchmark scores.
  Bold marks the best score within each backbone group.}
  \label{tab:temporal}
  \centering
  \setlength{\tabcolsep}{2.5pt}
  \small
  \begin{tabular}{@{}l ccc @{\hspace{8pt}} cccc@{}}
    \toprule
    \multirow{2}{*}{Model}
    & \multicolumn{3}{c}{Temporal Grounding}
    & \multicolumn{4}{c}{Complex Reasoning} \\
    \cmidrule(lr){2-4} \cmidrule(lr){5-8}
    & \makecell{QV\\Highlights}
    & \makecell{Activity\\Net}
    & \makecell{Charades\\STA}
    & \makecell{Video\\Holmes}
    & \makecell{Video\\MMMU}
    & \makecell{Video\\TT}
    & E.T.~Bench \\
    \midrule
    \multicolumn{8}{c}{Qwen3-VL-4B Backbone} \\
    \midrule
    Qwen3-VL-4B~\cite{qwen3vl2025}
    & 54.31 & 41.99 & 43.64 & 39.96 & 55.44 & 36.90 & 34.55 \\
    + SFT-Desc
    & 55.63 & 42.18 & 43.52 & 41.26 & 55.00 & \textbf{41.70} & 32.57 \\
    + SFT-DVC
    & 57.13 & 42.36 & 40.18 & \textbf{41.97} & \textbf{56.22} & 41.20 & 37.78 \\
    \textbf{+ EventLens}
    & \textbf{58.50} & \textbf{45.76} & \textbf{47.99}
    & 40.28 & 55.56 & 38.80 & \textbf{41.47} \\
    \midrule
    \multicolumn{8}{c}{Qwen3-VL-8B Backbone} \\
    \midrule
    Qwen3-VL-8B~\cite{qwen3vl2025}
    & 50.03 & 39.96 & 44.79 & 39.96 & \textbf{60.11} & 40.00 & 32.77 \\
    + SFT-Desc
    & 51.44 & 39.70 & 44.80 & 42.24 & 58.00 & 41.80 & 33.67 \\
    + SFT-DVC
    & 57.61 & 43.70 & 42.56 & \textbf{42.57} & 57.33 & \textbf{42.00} & \textbf{39.45} \\
    \textbf{+ EventLens}
    & \textbf{59.12} & \textbf{46.35} & \textbf{48.25}
    & 40.07 & 59.44 & 39.60 & 38.40 \\
    \bottomrule
  \end{tabular}
\end{table*}

\begin{table*}[t]
  \caption{General and long-video QA under the shared 256-frame protocol.
  Columns use official benchmark scores.
  Bold marks the best score within each backbone group.}
  \label{tab:general}
  \centering
  \setlength{\tabcolsep}{2.5pt}
  \small
  \begin{tabular}{@{}l cccc @{\hspace{8pt}} ccc@{}}
    \toprule
    \multirow{2}{*}{Model}
    & \multicolumn{4}{c}{General Video QA}
    & \multicolumn{3}{c}{Long Video QA} \\
    \cmidrule(lr){2-5} \cmidrule(lr){6-8}
    & \makecell{MV\\Bench} & \makecell{Temp\\Compass} & \makecell{AoT\\Bench} & \makecell{Vino\\Ground}
    & \makecell{Video\\MME} & MLVU & \makecell{LongVideo\\Bench} \\
    \midrule
    \multicolumn{8}{c}{Qwen3-VL-4B Backbone} \\
    \midrule
    Qwen3-VL-4B~\cite{qwen3vl2025}
    & 62.55 & 74.96 & 53.64 & 37.00 & 65.00 & 56.77 & 60.70 \\
    + SFT-Desc
    & 64.62 & \textbf{76.35} & \textbf{54.39} & 41.40 & 64.30 & 56.77 & 61.30 \\
    + SFT-DVC
    & \textbf{64.70} & 75.93 & 54.12 & 40.20 & 65.00 & 58.17 & \textbf{62.10} \\
    \textbf{+ EventLens}
    & 63.90 & 75.12 & 53.99 & \textbf{45.00}
    & \textbf{65.30} & \textbf{58.37} & 61.60 \\
    \midrule
    \multicolumn{8}{c}{Qwen3-VL-8B Backbone} \\
    \midrule
    Qwen3-VL-8B~\cite{qwen3vl2025}
    & 64.08 & 76.74 & 54.84 & 40.20 & 66.70 & \textbf{57.97} & 59.90 \\
    + SFT-Desc
    & \textbf{66.25} & 77.39 & 54.87 & 41.00 & 67.10 & 55.58 & 60.00 \\
    + SFT-DVC
    & 66.07 & \textbf{77.81} & \textbf{55.12} & 41.80 & 67.20 & \textbf{57.97} & 60.10 \\
    \textbf{+ EventLens}
    & 65.42 & 77.17 & 54.55 & \textbf{47.20}
    & \textbf{67.50} & 56.77 & \textbf{60.60} \\
    \bottomrule
  \end{tabular}
\end{table*}

\paragraph{EventLens delivers targeted temporal gains.}
Tables~\ref{tab:temporal} and~\ref{tab:general} show that EventLens is most reliable when
evaluation requires explicit event localization.
Across both Qwen3-VL backbones, it is the best variant in every temporal-grounding column, while
structured reasoning gains concentrate on event-sensitive benchmarks such as E.T.~Bench and
VinoGround.
On broad QA benchmarks, EventLens is not uniformly better than SFT baselines, which is
expected because these benchmarks mix temporal reasoning with caption-style semantics,
object recognition, and answer-format alignment.
The important observation is that EventLens preserves comparable broad-QA performance
while producing larger gains on the temporally targeted evaluations.

\FloatBarrier
\begin{table}[!htbp]
  \caption{Comparison with existing video post-training methods on Qwen2.5-VL-7B.
  Charades-STA and QVHighlights report mIoU; VinoGround reports the Text score and
  E.T. Bench the average score.}
  \label{tab:cr_same_backbone}
  \centering
  \small
  \setlength{\tabcolsep}{3pt}
  \begin{tabular}{@{}lrrrrrr@{}}
    \toprule
    Model & \makecell{Charades\\STA} & \makecell{QV\\Highlights}
    & \makecell{Vino\\Ground} & E.T. Bench & \makecell{MV\\Bench}
    & \makecell{LongVideo\\Bench} \\
    \midrule
    Qwen2.5-VL-7B~\cite{qwen25vl2025} & 38.34 & 27.66 & 29.40 & 30.96 & 63.38 & \textbf{59.20} \\
    Video-R1-7B~\cite{videor1_2025} & 28.80 & 22.33 & 31.20 & 29.50 & 60.27 & 56.20 \\
    \textbf{EventLens-7B} & \textbf{42.70} & \textbf{43.79} & \textbf{41.00} & \textbf{35.85} & \textbf{64.00} & 58.60 \\
    \bottomrule
  \end{tabular}
\end{table}

This pattern extends to the Qwen2.5-VL-7B comparison with other video post-training methods
(Table~\ref{tab:cr_same_backbone}).
The gains again concentrate on temporal grounding and event-sensitive reasoning, supporting
event-structure recovery as a transferable training objective across the two Qwen-VL generations.
Additional comparisons are reported in Appendix~\ref{app:cr_rl_comparisons}.

\subsection{Ablation studies}
\label{sec:ablation}

\begin{table}[!h]
  \caption{Teacher consolidation and task-composition ablation on Qwen3-VL-4B.
  R1--R3 denote segmentation, progression, and relation teachers.
  Full MT-OPD is EventLens with all three teachers; the single-teacher oracle selects
  the best individual score per metric.}
  \label{tab:ablation_teacher_consolidation}
  \centering
  \small
  \setlength{\tabcolsep}{2.5pt}
  \begin{tabular}{@{}>{\raggedright\arraybackslash}p{0.22\linewidth}cccccccc@{}}
    \toprule
    Variant
    & TimeLens
    & E.T.~Bench
    & \makecell{Video\\MME}
    & \makecell{Vino\\Ground}
    & \makecell{Video\\TT}
    & Holmes
    & MLVU
    & \makecell{LongVideo\\Bench} \\
    & avg mIoU & Avg & Avg & Score & Score & Acc & Avg & Avg \\
    \midrule
    Base & 46.65 & 34.55 & 65.00 & 37.00 & 36.90 & 39.96 & 56.77 & 60.70 \\
    \midrule
    R1 teacher & 50.84 & 40.34 & 65.30 & 39.20 & 38.20 & 40.12 & 58.76 & 61.10 \\
    R2 teacher & 50.76 & 40.94 & 65.40 & 45.40 & 39.50 & 40.61 & 59.76 & 61.80 \\
    R3 teacher & 50.96 & 39.87 & 65.00 & 37.80 & 37.20 & 39.19 & 57.57 & 61.60 \\
    Single-teacher oracle & 50.96 & 40.94 & 65.40 & 45.40 & 39.50 & 40.61 & 59.76 & 61.80 \\
    \midrule
    MT-OPD (R1+R2) & 50.73 & 41.44 & 65.40 & 45.00 & 39.50 & 40.72 & 59.36 & 61.70 \\
    Full MT-OPD & 50.75 & 41.47 & 65.30 & 45.00 & 38.80 & 40.28 & 58.37 & 61.60 \\
    \bottomrule
  \end{tabular}
\end{table}

\paragraph{Teacher specialization is complementary.}
Table~\ref{tab:ablation_teacher_consolidation} separates the task-specialized teachers before
they are compressed into one student.
The single-teacher rows do not reveal a universal best teacher: different skills lead on
target temporal grounding, retention, and reasoning metrics.
The teacher oracle is therefore an upper envelope, not a deployable policy.
MT-OPD recovers most of this envelope in one student, which supports the consolidation
objective.
Adding the relation teacher keeps temporal-grounding strength and slightly improves the
event-centric reasoning average, but does not dominate every retention metric.

\FloatBarrier
\begin{table}[!htbp]
  \centering
  \begin{minipage}[t]{0.50\linewidth}
  \vspace{0pt}
  \caption{Direct validation of recovery capabilities on Qwen3-VL-4B.
  R1--R3 denote segmentation, progression, and relation scores.}
  \label{tab:cr_task_validation}
  \centering
  \small
  \begin{tabular*}{\linewidth}{@{\extracolsep{\fill}}lrrr@{}}
    \toprule
    Variant & \makecell{R1\\Seg.} & \makecell{R2\\Prog.} & \makecell{R3\\Rel.} \\
    \midrule
    Base & 0.399 & 0.550 & 0.555 \\
    R1+R2 & 0.589 & 0.680 & 0.574 \\
    R1+R3 & 0.592 & 0.590 & 0.708 \\
    R2+R3 & 0.444 & 0.700 & 0.705 \\
    Full MT-OPD & 0.588 & 0.685 & 0.706 \\
    \bottomrule
  \end{tabular*}
  \end{minipage}
  \hfill
  \begin{minipage}[t]{0.46\linewidth}
  \vspace{0pt}
  \textbf{Direct validation separates the skills.}
  Omitting each recovery objective reduces its corresponding capability.
  Adding R3 to R1+R2 improves relation reasoning while preserving segmentation and progression, showing that inter-event dependencies complement boundary and intra-event skills.
  The full student stays within 0.015 of the best score in each column.
  MT-OPD thus retains complementary recovery capabilities in one model (individual teachers in Appendix~\ref{app:cr_task_validation}).
  \end{minipage}
\end{table}

\begin{figure*}[t]
  \centering
  \includegraphics[width=0.85\textwidth]{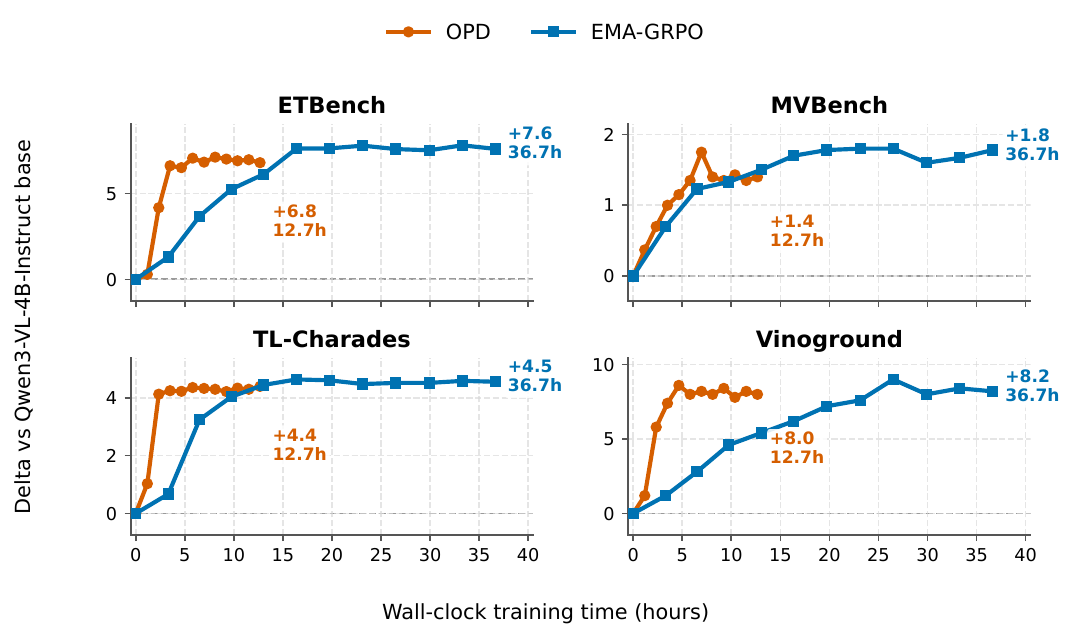}
  \vspace{-10pt}
  \caption{Wall-clock efficiency of EventLens versus direct Mixed-GRPO on four representative transfer
  metrics. The curves track MT-OPD student optimization and direct Mixed-GRPO training.}
  \label{fig:mt_opd_time_efficiency}
\end{figure*}

\FloatBarrier
\paragraph{MT-OPD mainly changes the optimization pathway.}
Direct Mixed-GRPO exposes one policy to heterogeneous rewards whose scales, sparsity patterns,
and output formats differ across segmentation, progression discrimination, and event-relation
reasoning.
EventLens instead lets each teacher first solve a cleaner task-specific RL problem, then
trains the student from dense top-$k$ token distributions on its own trajectories.
This converts sparse verifier outcomes into smoother token-level training signals,
providing a mechanism for the faster student-stage transfer in Fig.~\ref{fig:mt_opd_time_efficiency}: MT-OPD reaches a comparable
transfer regime after roughly 13 hours, while direct Mixed-GRPO needs roughly 37 hours.

\subsection{Analysis: Transfer and Failure Modes}
\label{sec:analysis_tradeoffs}

\paragraph{Why event-structure recovery transfers.}
EventLens transfers across tasks because its proxy supervision does not teach a particular
downstream output format; it teaches the model to bind textual decisions to visually grounded
temporal units.
Segmentation, progression, and relation tasks use interval lists, direction judgments, and
choices or permutations, yet transfer to timestamp localization, event-level QA, and long-video QA.
The common operation is boundary-sensitive temporal comparison: the model must identify when
an event starts and ends, decide whether a state transition is plausible, and select the
neighboring event that best preserves the visual process.
This inductive bias explains why gains are most stable on temporal grounding and
event-sensitive reasoning, where success depends on localized temporal evidence.

\begin{table}[t]
  \caption{Robustness to timestamp perturbation in event supervision on Qwen3-VL-4B.
  Training uses 100 GRPO steps and 2,000 samples per recovery family; percentages denote perturbed chains.
  VinoGround reports the Group score.}
  \label{tab:cr_jitter_main}
  \centering
  \small
  \setlength{\tabcolsep}{2.5pt}
  \begin{tabular}{@{}lrrrrrr@{}}
    \toprule
    \makecell{Training\\annotations} & \makecell{VinoGround\\Group} & MVBench & \makecell{LongVideo\\Bench} & \makecell{Charades\\mIoU} & \makecell{QVHighlights\\mIoU} & \makecell{E.T. Bench\\Avg.} \\
    \midrule
    Instruction base & 37.00 & 62.55 & 60.70 & 43.64 & 54.31 & 34.55 \\
    Clean & 39.00 & 63.38 & 61.20 & 45.70 & 58.16 & 38.20 \\
    10\% jitter & 38.60 & 63.20 & 61.30 & 45.61 & 57.99 & 37.92 \\
    20\% jitter & 38.60 & 63.40 & 61.00 & 45.42 & 57.58 & 37.71 \\
    \bottomrule
  \end{tabular}
\end{table}

\FloatBarrier

\paragraph{Supervision robustness.}
EventLens remains robust to moderate perturbations of the automatically reconstructed temporal supervision. With 20\% perturbation, all six metrics remain above the instruction baseline and close to clean training (Table~\ref{tab:cr_jitter_main}), suggesting that the model learns event-level temporal structure rather than relying on exact boundary annotations.

\paragraph{Failure modes and trade-offs.}
In temporal grounding, EventLens can still confuse adjacent events when visual states change
gradually, or when a query describes a high-level activity spanning several fine-grained
actions.
In multi-event relation tasks, visually similar repeated actions remain difficult because the
correct answer may depend on subtle context rather than the appearance of an individual clip.
These cases suggest that EventLens improves sensitivity to event boundaries and local temporal
dependencies, but does not solve all long-range narrative reasoning.
Appendix~\ref{app:cr_capability_subsets} reports the complete capability-specific results
alongside qualitative cases.

\section{Conclusion}
\label{sec:conclusion}
We present \emph{EventLens}, a vision-centric reinforcement learning framework for structured temporal video understanding.
Instead of learning from caption-style supervision alone, EventLens trains models to recover event structure from visual evidence, including event boundaries, progression, and inter-event relations. 
By combining verifiable event-structure recovery tasks with task-specialized RL and policy consolidation, EventLens achieves consistent gains on temporal grounding and structured temporal reasoning while preserving broad video understanding ability. 
More broadly, our results suggest that recovering event structure from visual evidence provides an effective training principle for advancing temporal understanding in video MLLMs, beyond treating temporal reasoning as a byproduct of language supervision.

\begin{ack}
This work was supported in part by the Beijing Major Science and Technology Project under Contract no.  Z251100008425023.

\end{ack}

\bibliographystyle{abbrvnat-nourl}
\bibliography{references}

\newpage
\appendix
\section*{Appendix}
\raggedbottom

This appendix provides additional implementation details, evaluation protocols, and qualitative analyses that complement the main paper.
We first describe the construction of the event-structure proxy-task pool, including data sources, recovery views, reward parsing rules, and the multi-teacher OPD implementation (\S\ref{app:data_details}--\ref{app:opd_impl}).
We then report training hyperparameters and the detailed temporal grounding evaluation protocol for reproducibility (\S\ref{app:hyperparams}--\ref{app:temporal_grounding_eval}).
Next, we provide additional empirical analyses, including comparisons with related video post-training methods, qualitative case studies, and full efficiency curves beyond the representative metrics shown in the main text (\S\ref{app:videossr_peer}--\ref{app:opd_efficiency_full_suite}).
Finally, we discuss the current limitations of EventLens and directions for future work (\S\ref{sec:limitations}).

\makeatletter
\newcommand{\appnonfloatcaption}[2]{%
  \refstepcounter{figure}%
  \label{#1}%
  \@makecaption{\fnum@figure}{#2}%
}
\newcommand{\appnonfloattablecaption}[2]{%
  \refstepcounter{table}%
  \label{#1}%
  {\setlength{\abovecaptionskip}{0pt}%
   \setlength{\belowcaptionskip}{7pt}%
   \@makecaption{\fnum@table}{#2}}%
}
\makeatother

\section{Proxy-task data details}
\label{app:data_details}

The post-training pool converts source videos into three event-structure recovery views
that mirror the main EventLens tasks.
Table~\ref{tab:appendix_proxy_data} summarizes what each view asks the model to recover
and which deterministic target is used for verification.
All instances are built from the official training partitions of
TimeLens-100K~\cite{timelens2025} and ET-Instruct-164K~\cite{etbench2024};
videos and annotations from held-out test splits of TimeLens, E.T.~Bench, or any other
evaluation benchmark are excluded from the post-training pool.
Thus, naming overlap between a training source and an evaluation benchmark does not imply
train--test sharing.
For reproducibility, each stored example contains a prompt, one or more video frame lists,
a structured target, metadata, and a recovery-family label used only to select the matching
verifier and teacher.

\begin{table}[h]
  \caption{Event-structure recovery views used for post-training.}
  \label{tab:appendix_proxy_data}
  \centering
  \footnotesize
  \setlength{\tabcolsep}{3pt}
  \begin{tabular}{@{}p{0.22\linewidth} p{0.49\linewidth} p{0.23\linewidth}@{}}
    \toprule
    Recovery view & Constructed recovery task & Deterministic target \\
    \midrule
    Event-boundary recovery
    & Recover phase-, event-, or action-level time spans requested by the prompt.
    & Event intervals; balanced by level and domain \\
    Single-event progression
    & Decide whether an event-level clip is played forward or backward.
    & Direction label; balanced by playback direction and domain \\
    Multi-event relation
    & Recover the next event, a missing middle event, or the order of shuffled clips.
    & Choice or ordering; balanced by relation subtask and domain \\
    \bottomrule
  \end{tabular}
\end{table}

\begin{wrapfigure}{r}{0.46\linewidth}
  \vspace{-0.8em}
  \centering
  \includegraphics[width=\linewidth]{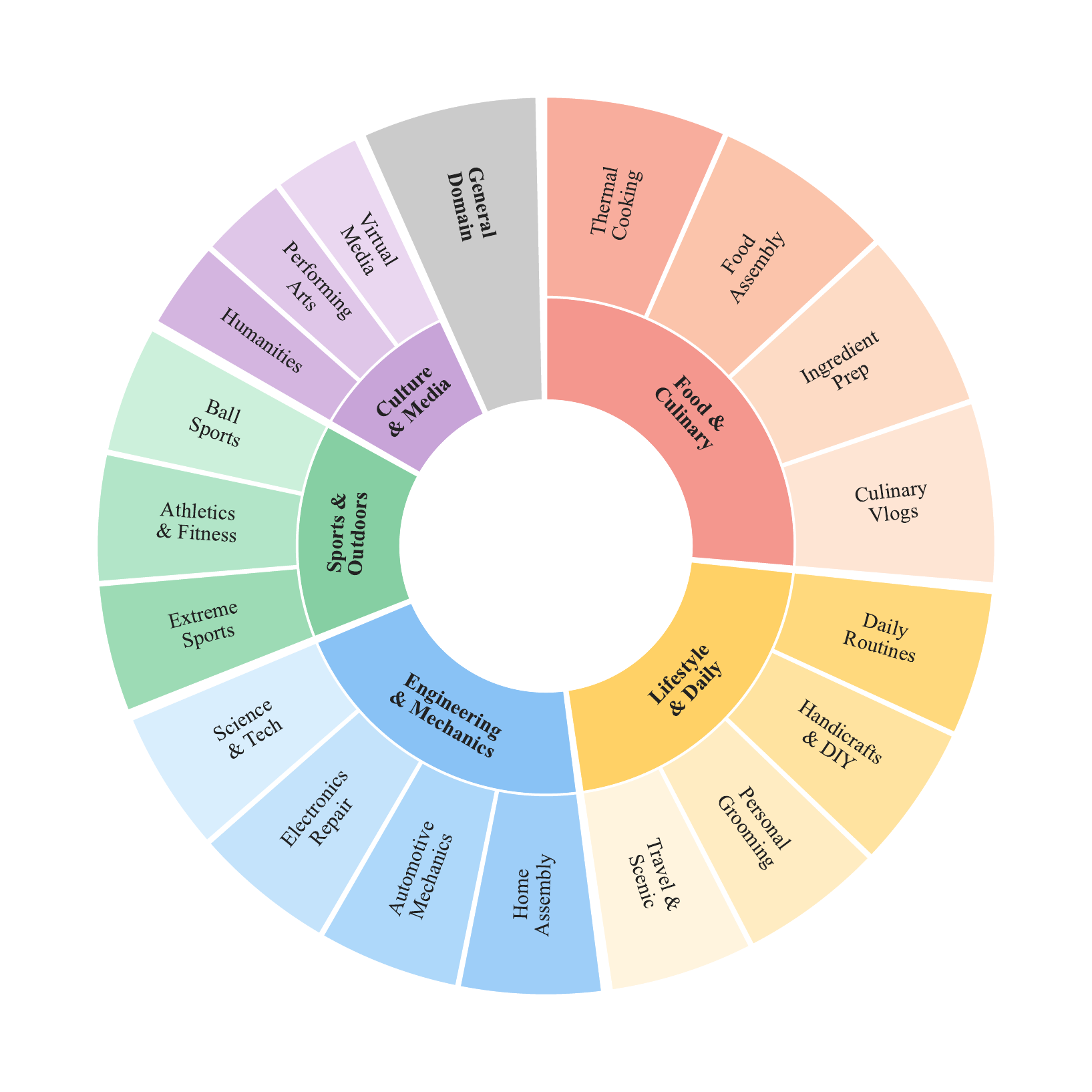}
  \vspace{-0.8em}
  \caption{
  Domain composition of the input videos used to construct proxy-task instances.
  }
  \label{fig:app_input_domain_sunburst}
  \vspace{-0.7em}
\end{wrapfigure}

The proxy-task pool is intentionally built from a diverse set of video domains rather than
from a single narrow source.
Fig.~\ref{fig:app_input_domain_sunburst} visualizes the domain composition of the input
videos before conversion into segmentation, single-event progression, and multi-event relation instances.
This diversity is important because the proxy tasks manipulate event structure rather than
dataset-specific labels: the same verifier design can supervise event boundaries,
single-event progression, and multi-event relations across instructional, human-activity,
object-centric, and scene-level videos.

For video inputs, we use offline frame lists rather than decoding videos during training.
The experiment builder derives frame lists from trusted 2 FPS frame caches and applies a
duration-dependent frame policy: clips shorter than 60 seconds are kept at 2 FPS, while longer
clips are sampled at 1 FPS, followed by a uniform cap of 256 frames.
The same 256-frame and $256{\times}256$ pixel budget is used for training and evaluation.

\subsection{Hierarchy construction and supervision assessment}
\label{app:cr_data_quality}
Figure~\ref{fig:cr_construction} uses a curling news report to illustrate how candidate
shots are organized into a temporal hierarchy. The sampled frames span a studio
presentation, the curling-club exterior, on-ice demonstrations, and player
interviews. Camera changes supply candidate boundaries; semantic refinement
determines which consecutive or within-shot intervals form coherent L2 events.
\begin{center}
\begin{minipage}{\linewidth}
\appnonfloattablecaption{tab:cr_construction_examples}{Semantic refinement and temporal hierarchy in the curling-report schematic.
Frame identifiers refer to Figure~\ref{fig:cr_construction}; event boundaries and assignments are illustrative.}
  \centering
  \small
  \setlength{\tabcolsep}{3pt}
  \renewcommand{\arraystretch}{1.12}
  \begin{tabular}{@{}>{\raggedright\arraybackslash}p{0.12\linewidth}>{\raggedright\arraybackslash}p{0.38\linewidth}>{\raggedright\arraybackslash}p{0.45\linewidth}@{}}
    \toprule
    Decision & Report content & Semantic interpretation \\
    \midrule
    KEEP & Interview view $c_5$: one speaker continues within a single shot.
    & Retain the candidate interval as one L2 event when its semantic focus remains unchanged. \\
    \addlinespace[3pt]
    MERGE & Adjacent views $c_3$ and $c_4$ show the same group performing a curling demonstration.
    & Combine consecutive candidates $c_i$ and $c_{i+1}$ into one L2 event when a camera cut preserves the activity. \\
    \addlinespace[3pt]
    SPLIT & A single rink shot can cover successive segments, such as preparing a delivery and demonstrating the play.
    & Place an internal boundary at a change of semantic focus, producing separate L2 events $e_i$ and $e_{i+1}$. \\
    \midrule
    L1 phase & The complete curling report, from the studio presentation to the return to the anchors.
    & Group related L2 events under the shared topic, ``Curling report,'' while preserving their temporal order. \\
    \addlinespace[3pt]
    L2 event & Club context, on-ice demonstrations, and player interviews provide distinct event-level content.
    & Describe each coherent report segment with an event interval; $e_1$, $e_2$, and $e_3$ illustrate the event sequence. \\
    \addlinespace[3pt]
    L3 action & Local actions, such as releasing a stone or sweeping, occur inside an on-ice demonstration.
    & Keep each action interval inside its parent L2 event, and each L2 event inside the enclosing L1 phase. \\
    \bottomrule
  \end{tabular}
\end{minipage}
\end{center}
\paragraph{Temporal organization.}
The hierarchy separates a report's overall topic, its coherent semantic segments,
and the local actions within each segment. A camera cut can occur inside one event,
while a single shot can span an event transition. Table~\ref{tab:cr_construction_examples}
therefore ties refinement to continuity or a change of semantic focus. Once event
intervals are established, L3 actions remain temporally contained in their parent
events, and related L2 events are aggregated into an L1 phase.
\paragraph{Construction implementation.}
We sample annotation frames at 1 FPS and use TransNetV2 to obtain candidate shot intervals.
Semantic refinement retains a shot as one event when its focus remains unchanged,
merges consecutive shots that preserve event continuity, and splits a shot at semantic transitions.
The resulting L2 events contain L3 action units that capture atomic motion or state transitions;
adjacent L2 events sharing a high-level intent, topic, or discourse function are grouped into L1 phases.
Before task conversion, we check scene coverage, temporal validity, parent--child containment,
temporal ordering, and non-overlap. Extremely short, ambiguous, or overly fragmented intervals
are discarded. Each segmentation prompt requests one temporal level, and progression clips
preserve coherent event-level state transitions. For relation recovery, distractors are sampled
preferentially from the same video or semantically related neighboring contexts; candidates are
visual clips with neutral identifiers rather than textual descriptions.
\par
\paragraph{Model-assisted audit.}
We inspect 100 sampled event chains using three MLLM judges and four criteria: caption grounding, boundary acceptability, hierarchy consistency, and proxy answerability. Table~\ref{tab:cr_quality} reports per-judge and majority-voted acceptance; 87\% of chains satisfy all four criteria under majority voting.
\begin{center}
\begin{minipage}{\linewidth}
\appnonfloattablecaption{tab:cr_quality}{Model-assisted quality audit of 100 event chains. Columns report acceptance for each criterion and for all four criteria jointly.}
  \centering
  \footnotesize
  \setlength{\tabcolsep}{3pt}
  \begin{tabular}{@{}lrrrrr@{}}
    \toprule
    Judge & \makecell{Caption\\grounded} & \makecell{Boundary\\acceptable} & \makecell{Hierarchy\\consistent} & \makecell{Proxy\\answerable} & \makecell{All four\\passed} \\
    \midrule
    GPT-5.5 & 97\% & 91\% & 97\% & 87\% & 83\% \\
    Claude-4.8 & 94\% & 89\% & 95\% & 88\% & 73\% \\
    Gemini-2.5-Pro & 99\% & 97\% & 100\% & 82\% & 82\% \\
    Three-judge majority & 98\% & 93\% & 98\% & 91\% & 87\% \\
    \bottomrule
  \end{tabular}
\end{minipage}
\end{center}
\paragraph{Timestamp perturbation.}
Table~\ref{tab:cr_jitter_main} reports the Qwen3-VL-4B experiment using 2,000 samples per recovery family and 100 GRPO steps. We select 10\% or 20\% of chains and perturb their temporal coordinates while preserving interval duration, event text, and sample identity. Percentages denote the selected-chain fraction, not boundary displacement magnitude. This protocol assesses timestamp imprecision in recovery supervision.

\section{Reward and parsing details}
\label{app:reward_details}

The reward parser follows the same three event-structure recovery families used in
Section~\ref{sec:proxy}: event-boundary recovery, single-event progression, and multi-event
relations.
This section specifies how model text is converted into the verifier inputs in
Eqs.~\ref{eq:seg_reward} and~\ref{eq:event_reward}.
Format validity is logged for analysis, but the training reward is always the task verifier
defined by the constructed target structure.

\paragraph{Event-boundary recovery.}
Segmentation responses are parsed from the \texttt{<events>} block as a list of
$[s,e]$ intervals.
Invalid intervals, missing tags, malformed bracket formats, and repeated event tags receive
zero reward because no reliable boundary set can be recovered.
Valid predictions are de-duplicated with one-dimensional NMS and matched to ground-truth
intervals with Hungarian matching.
The matched intervals are then scored by the F1-IoU reward in Eq.~\ref{eq:seg_reward},
so the same parser supports phase-, event-, and action-level segmentation prompts.

\paragraph{Single-event progression.}
Progression-discrimination responses are parsed as a multiple-choice decision over the
forward and reverse playback options.
The parser first extracts the final letter inside \texttt{<answer>} and falls back to the
last standalone option letter if the tag is missing.
The reward is one only when the parsed option matches the constructed playback direction
and zero otherwise, matching $R_{\text{prog}}$ in Section~\ref{sec:proxy}.
\paragraph{Multi-event relations.}
Next-event prediction and fill-in use the same multiple-choice parser as progression
discrimination, but their target option is derived from the hidden event in the local
event window.
Clip reordering instead parses a numeric permutation from \texttt{<answer>}.
Responses with no parseable in-range clip index receive zero reward.
For parseable but incomplete permutations, duplicate indices are removed, overlong
predictions are truncated to the target length, and missing target indices are appended
before scoring so that every valid partial order maps to a complete permutation.
We then compute the normalized temporal-jigsaw displacement in Eq.~\ref{eq:event_reward},
which gives full credit to the exact order and partial credit when clips remain near their
correct temporal positions.

\section{Multi-teacher OPD implementation}
\label{app:opd_impl}

The implementation of multi-teacher OPD follows a simple routed distillation loop.
Unlike GRPO teacher training, OPD uses one student rollout per prompt and does not compute
reward groups, advantages, online filtering, or critic values.
After the student generates an on-policy response, the corresponding teacher scores the same
student trajectory with top-$k$ next-token log probabilities.
The student then gathers its log probabilities at the teacher top-$k$ token ids and minimizes
a sparse forward KL after renormalizing both distributions on this support.

\begin{enumerate}
  \item Sample prompts from the mixed proxy-task pool.
  \item Generate responses $y \sim \pi_\theta(\cdot\mid x)$ with the current student.
  \item Route each example to the teacher trained for its recovery family.
  \item Query the routed teacher for top-$k$ token distributions along $y$.
  \item Update the student with the sparse OPD KL loss.
\end{enumerate}

Teacher routing follows the same task grouping as the data pipeline:
segmentation examples are routed to the segmentation teacher,
single-event progression examples to the progression teacher,
and multi-event relation examples to the relation teacher.
For memory efficiency, teacher models are treated as reference modules and can be loaded,
queried on their sub-batch, and offloaded sequentially.
We also use task-homogeneous batching so that each mini-batch contains examples from a single
coarse task group whenever possible.

\section{Training hyperparameters}
\label{app:hyperparams}

Appendix~\ref{app:hyperparams} reports the shared implementation settings used across
all Qwen3-VL backbones unless otherwise specified.
The teacher-related rows correspond to task-specialized EMA-GRPO training, while the
OPD-related rows correspond to the final multi-teacher student consolidation stage.
We keep the video budget and decoding protocol fixed across pretrained, SFT, teacher,
and EventLens variants, so that reported differences are attributable to post-training
rather than inference-time frame allocation, resolution, or sampling choices.

\begin{center}
\begin{minipage}{0.78\textwidth}
  \appnonfloattablecaption{tab:appendix_hparams}{Training and inference settings.}
  \vspace{0.2em}
  \footnotesize
  \setlength{\tabcolsep}{0pt}
  \begin{tabular*}{\linewidth}{@{\extracolsep{\fill}}llll@{}}
    \toprule
    Item & Value & Item & Value \\
    \midrule
    Backbones & Qwen3-VL-4B / 8B
    & Teacher training & EMA-GRPO \\
    Student training & Multi-teacher OPD
    & Teacher LR & $5 \times 10^{-7}$ \\
    Teacher batch & 64
    & GRPO rollouts & $G=8$ \\
    Teacher KL & $1 \times 10^{-2}$
    & OPD rollouts & $n=1$ \\
    OPD teacher dist. & top-$k=10$
    & OPD KL & 1.0 \\
    Teacher routing & By recovery family
    & Eval. decoding & Greedy \\
    Max frames & 256
    & Max pixels & $256{\times}256$ \\
    \bottomrule
  \end{tabular*}
\end{minipage}
\end{center}

The one-rollout OPD setting is used because MT-OPD obtains token-level teacher
distributions on the student trajectory, rather than estimating group-normalized rewards
as in GRPO.
The top-$k$ teacher distribution keeps the routed multi-teacher distillation target
sparse and memory-efficient while preserving high-probability alternatives for stable
KL matching.

\section{Individual-teacher recovery capabilities}
\label{app:cr_task_validation}
\begin{center}
\begin{minipage}{\linewidth}
\appnonfloattablecaption{tab:cr_full_task_validation}{Individual-teacher recovery scores on Qwen3-VL-4B.}
  \centering
  \small
  \setlength{\tabcolsep}{3pt}
  \begin{tabular}{@{}lrrr@{}}
    \toprule
    Variant & \makecell{R1\\Segmentation} & \makecell{R2\\Progression} & \makecell{R3\\Relation} \\
    \midrule
    Base & 0.399 & 0.550 & 0.555 \\
    R1 & 0.587 & 0.570 & 0.580 \\
    R2 & 0.435 & 0.685 & 0.570 \\
    R3 & 0.450 & 0.565 & 0.705 \\
    Full MT-OPD & 0.588 & 0.685 & 0.706 \\
    \bottomrule
  \end{tabular}
\end{minipage}
\end{center}

\paragraph{Specialization and consolidation.}
Each teacher produces its largest gain on its corresponding recovery task.
Relative to Base, R1 improves segmentation by 0.188, R2 improves progression by 0.135,
and R3 improves relation reasoning by 0.150; their gains on the other two tasks are smaller.
Full MT-OPD matches or exceeds the best individual teacher in every column.
Together with the task-combination results in Table~\ref{tab:cr_task_validation},
these results support training complementary teacher policies and consolidating their
recovery capabilities into a single student.

\section{Temporal grounding evaluation details}
\label{app:temporal_grounding_eval}

We evaluate temporal grounding with the same VLMEval-style implementation for all
pretrained, SFT, and EventLens variants.
The three reported TimeLens subsets, QVHighlights, ActivityNet, and Charades-STA, are
loaded through the \texttt{TimeLensBench} interface, which converts each annotation into
a TSV row containing the video id, natural-language query, ground-truth span, and video
duration.
Videos are supplied as structured video inputs when the evaluated backend supports video
messages; otherwise the framework extracts frames with the same adaptive policy used in
the rest of the paper: videos up to 60 seconds are sampled at 2 FPS, videos between 60 and
256 seconds are sampled at 1 FPS, and longer videos are uniformly capped at 256 frames.
All models are decoded greedily, and no benchmark-specific demonstrations or external
judges are used for these temporal grounding scores.

For the main temporal grounding results, we use the following direct prompt:

\begin{quote}\small
Please find the visual event described by a sentence in the video, determining its starting
and ending times. The format should be: ``The event happens in the start time - end time
seconds''. For example: The event ``person turn a light on'' happens in the 24.3 - 30.4
seconds. Now I will give you the textual sentence: \texttt{\{query\}}. Please return its
start time and end time in seconds.
\end{quote}

The evaluation parser extracts the first valid pair of decimal timestamps from model
outputs in either ``$X$--$Y$'' or ``$X$ to $Y$'' form and discards pairs whose end time is
earlier than the start time.
For a predicted span $[\hat{s},\hat{e}]$ and a reference span $[s,e]$, temporal IoU is
computed as
\begin{equation}
\mathrm{tIoU}
= \frac{\max(0, \min(\hat{e}, e)-\max(\hat{s}, s))}
        {\max(\hat{e}, e)-\min(\hat{s}, s)} .
\end{equation}
If no valid timestamp pair is found, the example receives zero IoU; malformed reference
annotations are excluded from aggregation.
We report mean IoU (mIoU) in the main tables and additionally log recall at
IoU thresholds 0.3, 0.5, and 0.7 for reproducibility.

\section{Comparison with related video post-training work}
\label{app:videossr_peer}
\label{app:cr_rl_comparisons}

\begin{center}
\begin{minipage}{\linewidth}
\appnonfloattablecaption{tab:cr_full_comparison}{Comparisons with related video post-training methods. Charades-STA, QVHighlights, and ActivityNet report mIoU; VinoGround reports the Text score and E.T. Bench the average score. Bold marks the best score per column.}
  \vspace{0.2em}
  \centering
  \footnotesize
  \setlength{\tabcolsep}{2.2pt}
  \begin{tabular}{@{}lrrrrrrr@{}}
    \toprule
    Model & \makecell{Charades\\STA} & \makecell{QV\\Highlights} & \makecell{Activity\\Net} & \makecell{Vino\\Ground Text} & \makecell{E.T. Bench\\Avg.} & \makecell{MV\\Bench} & \makecell{LongVideo\\Bench} \\
    \midrule
    Qwen3-VL-8B~\cite{qwen3vl2025} & 44.79 & 50.03 & 39.96 & 40.20 & 32.77 & 64.08 & 59.90 \\
    VideoSSR-8B~\cite{videossr2026} & 41.52 & 53.67 & 39.81 & \textbf{49.20} & \textbf{41.94} & 66.72 & 60.90 \\
    VideoAuto-R1-8B~\cite{videoautor1_2026} & 41.95 & 56.94 & 42.54 & 43.60 & 39.36 & \textbf{67.07} & \textbf{62.70} \\
    OneThinker-8B~\cite{onethinker2025} & 40.09 & 54.07 & 38.90 & 36.80 & 31.14 & 59.77 & 57.80 \\
    \midrule
    EventLens-4B & 47.99 & 58.50 & 45.76 & 45.00 & 41.47 & 63.90 & 61.60 \\
    EventLens-8B & \textbf{48.25} & \textbf{59.12} & \textbf{46.35} & 47.20 & 38.40 & 65.42 & 60.60 \\
    \bottomrule
  \end{tabular}
\end{minipage}
\end{center}

Table~\ref{tab:cr_same_backbone} reports the Qwen2.5-VL-7B comparison in the main text.
Table~\ref{tab:cr_full_comparison} brings together the additional video post-training methods,
with each model listed once across the reported benchmarks.
EventLens-8B achieves the highest reported scores on Charades-STA, QVHighlights, and ActivityNet,
connecting event-structure recovery to temporal grounding.
Against VideoSSR-8B, EventLens improves the mean TimeLens score from 45.00 to 51.24.

\section{Qualitative case study}
\label{app:qualitative_case}

\par\medskip
\noindent\begin{minipage}{\textwidth}
  {\centering
  \includegraphics[width=\textwidth]{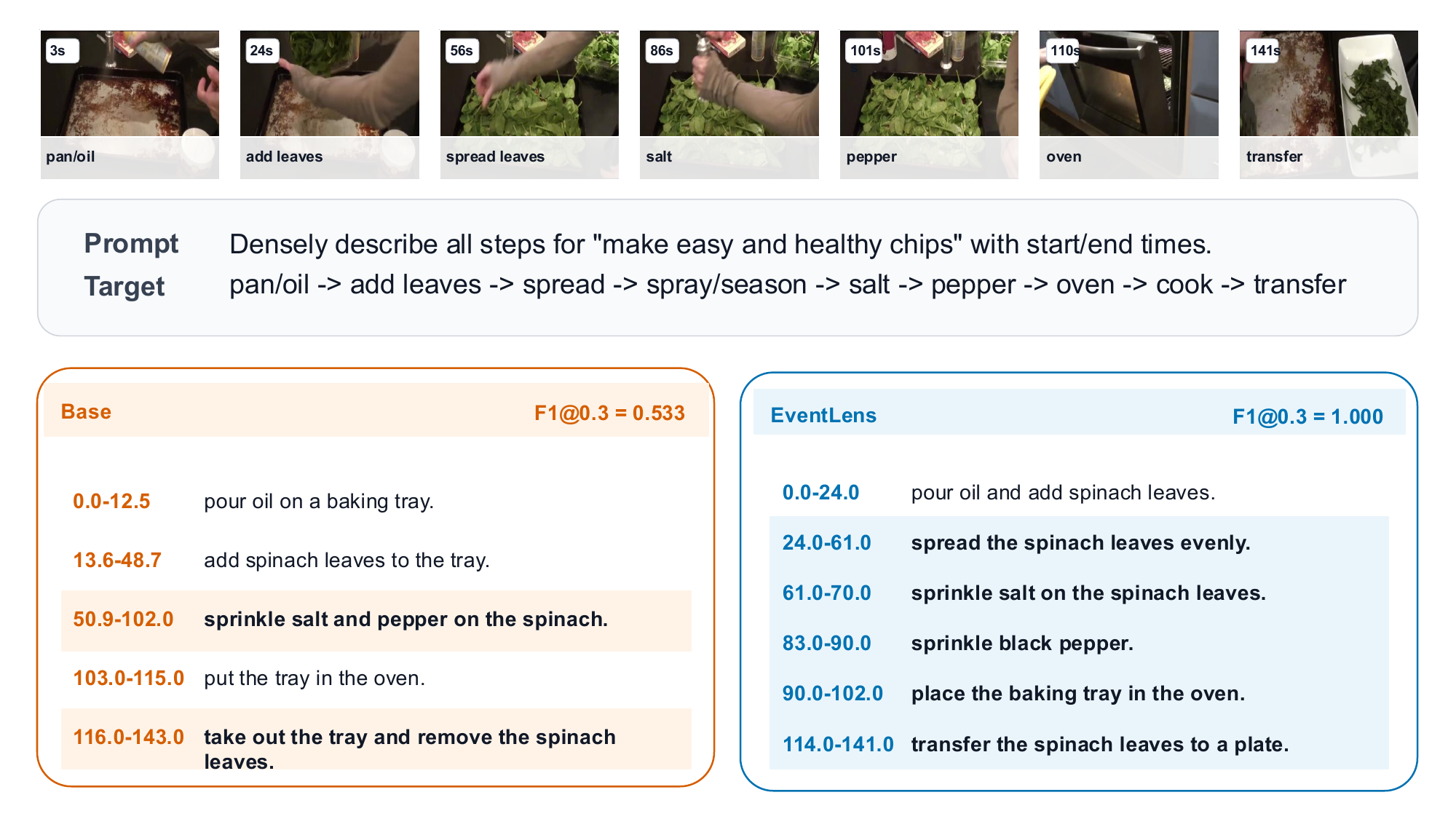}\par}
  \appnonfloatcaption{fig:app_etbench_healthy_chips_case}{Qualitative E.T.~Bench case on the instruction ``make easy and healthy chips.''
  The base model recovers the coarse cooking procedure but merges several adjacent
  fine-grained events, while EventLens trained with MT-OPD separates the event
  boundaries more faithfully and obtains perfect F1@0.3 on this example.}
\end{minipage}
\par\medskip

Fig.~\ref{fig:app_etbench_healthy_chips_case} illustrates the difference between coarse
procedure recognition and event-structure recovery.
The base model identifies the overall recipe flow, but its segments remain too coarse:
it merges the salt and pepper operations into one long interval and shifts the final
transfer step toward the oven-removal phase.
These errors are understandable from a captioning perspective because the predicted steps
are semantically plausible, but they are harmful for dense event evaluation where adjacent
visual actions must be separated in time.

EventLens with MT-OPD produces a more temporally factorized description.
It groups the initial oiling and leaf placement when the visual evidence is continuous,
then separates spreading, salting, peppering, oven placement, and transfer into distinct
intervals.
This case supports the main quantitative trend: the benefit of EventLens is not merely
better recipe semantics, but improved alignment between textual event descriptions and
their visual temporal boundaries.
It also clarifies why some caption-style SFT baselines can remain competitive on broad
reasoning scores while still missing the structured boundary precision targeted by our
proxy tasks.

\begin{figure}[!htbp]
  \centering
  \includegraphics[width=\textwidth]{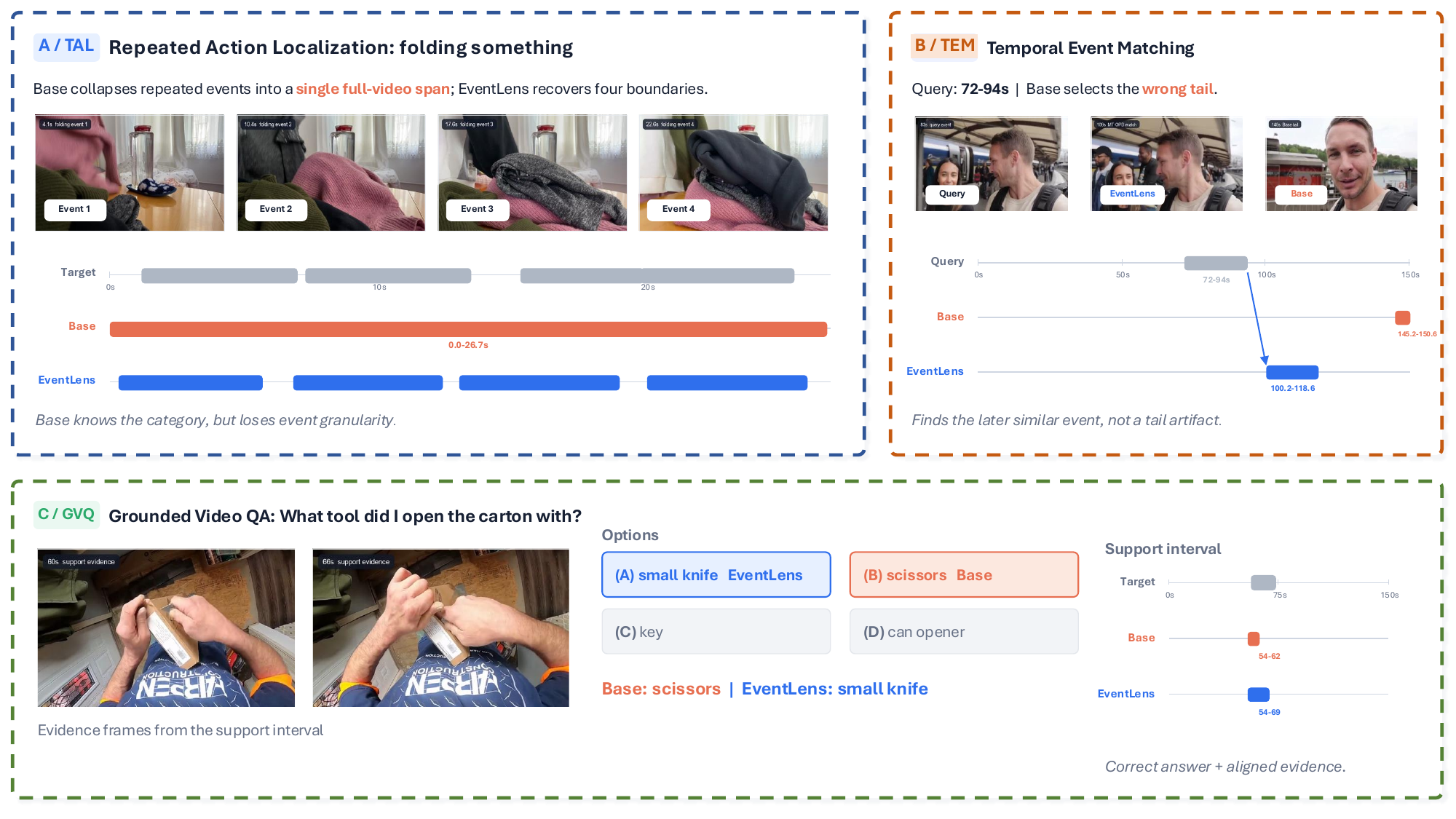}
  \caption{Additional E.T.~Bench qualitative cases covering repeated action localization,
  temporal event matching, and grounded video QA.
  Across the three settings, the base model often recognizes the coarse visual category
  or plausible answer but misaligns it with the relevant temporal evidence, while
  EventLens with MT-OPD preserves finer event boundaries and support intervals.}
  \label{fig:app_etbench_temporal_reasoning_cases}
\end{figure}

Fig.~\ref{fig:app_etbench_temporal_reasoning_cases} shows that the qualitative behavior
of EventLens extends beyond a single dense procedure example.
In repeated action localization, the base model collapses several visually similar
``folding'' occurrences into one nearly full-video prediction, whereas MT-OPD recovers
the repeated event granularity.
In temporal event matching, the base model drifts to a tail artifact, while MT-OPD
selects the later visually matched event segment.
In grounded video QA, the base model chooses a plausible but incorrect tool answer,
whereas MT-OPD selects the small knife and aligns the answer with the supporting frames.
These examples support the role of event-structure supervision: it improves not only
whether the model names the right event, but also whether the answer is grounded in the
correct temporal evidence.

\FloatBarrier
\section{Capability-specific transfer and trade-offs}
\label{app:cr_capability_subsets}
\begin{center}
\begin{minipage}{\linewidth}
\appnonfloattablecaption{tab:cr_capability_subsets}{Capability-level changes of EventLens relative to the corresponding Qwen3-VL instruction base. $\Delta$ denotes score-point differences for 4B and 8B models. The LongVideoBench entry retains the reported `Overall / TOS' label and is separate from the aggregate score in Table~\ref{tab:general}.}
  \centering
  \footnotesize
  \setlength{\tabcolsep}{3pt}
  \begin{tabular}{@{}>{\raggedright\arraybackslash}p{0.24\linewidth}>{\raggedright\arraybackslash}p{0.43\linewidth}rr@{}}
    \toprule
    Structural capability & Benchmark subset & \makecell{$\Delta$\\4B} & \makecell{$\Delta$\\8B} \\
    \midrule
    Event localization & E.T. Bench TAL/F1 & +7.61 & +7.87 \\
    Event grounding & E.T. Bench GND/F1 & +5.14 & +8.33 \\
    Temporal order & MLVU Order & +8.57 & +1.43 \\
    Motion direction & TempCompass Direction & +1.06 & +1.36 \\
    Text-event matching & VinoGround Text Score & +8.00 & +7.00 \\
    Group-level matching & VinoGround Group Score & +2.40 & +1.80 \\
    Long-range temporal ordering & LongVideoBench Overall / TOS & -4.10 & -1.40 \\
    Montage plot attributes & Video-TT Plot Attribute (Montage) & -4.08 & -4.08 \\
    Motion speed & TempCompass Speed & -2.84 & -2.21 \\
    Anomaly recognition & MLVU Anomaly Recognition & -2.56 & -2.56 \\
    \bottomrule
  \end{tabular}
\end{minipage}
\end{center}

\paragraph{Transfer aligns with the recovery objectives.}
Table~\ref{tab:cr_capability_subsets} shows gains on event-sensitive capabilities at both model scales.
Event localization improves by 7.61/7.87 points at 4B/8B, event grounding by 5.14/8.33 points,
and text-event matching by 8.00/7.00 points.
Temporal ordering, motion direction, and group-level matching also improve at both scales.
These gains align with the boundary localization, intra-event progression, and inter-event
relations emphasized by the three recovery objectives.

\paragraph{Subsets probe different information.}
Broad video-QA benchmarks combine temporal reasoning with static recognition, scene semantics,
external knowledge, and answer-format compliance.
The negative changes on montage plot attributes, motion-speed estimation, and anomaly recognition
occur at both scales; these subsets additionally probe global plot organization, motion rates,
and atypical event semantics.
Narrative-scale ordering also calls for context beyond the finite event windows used by the
current relation objective.
This capability profile motivates complementing event-structure recovery with narrative-scale
and fine-grained perceptual objectives.

\section{Additional EventLens efficiency curves}
\label{app:opd_efficiency_full_suite}

\begin{wrapfigure}{r}{0.58\linewidth}
  \vspace{-1.2em}
  \centering
  \includegraphics[width=\linewidth]{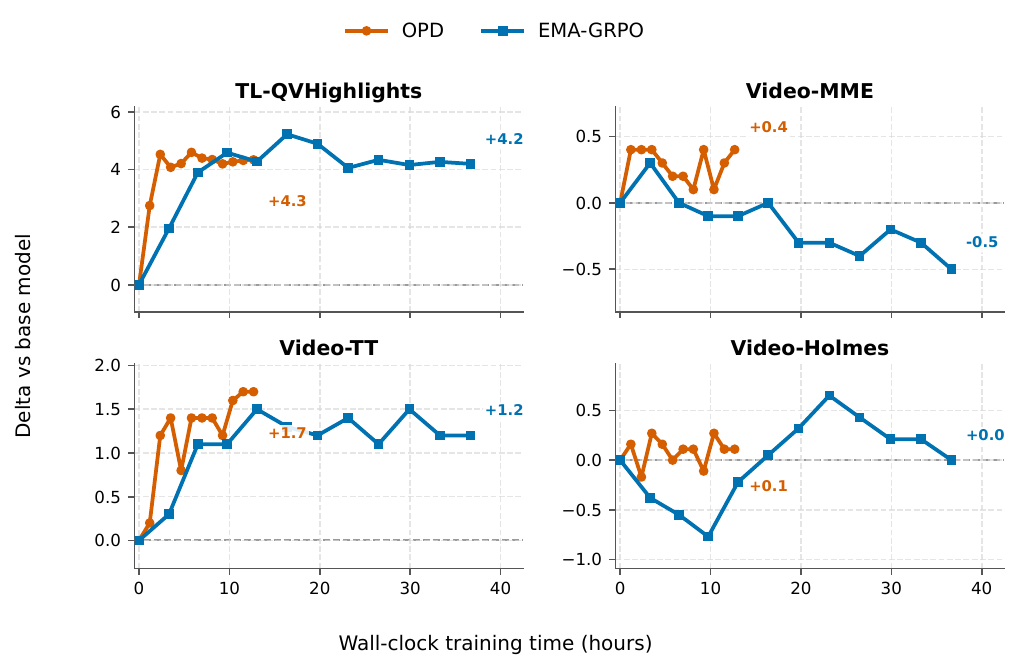}
  \vspace{-0.9em}
  \caption{
  Additional wall-clock transfer curves complementing the four representative metrics in
  Fig.~\ref{fig:mt_opd_time_efficiency}.
  }
  \label{fig:app_mt_opd_extra_curves}
  \vspace{-0.6em}
\end{wrapfigure}

Section~\ref{sec:exp} reports four representative transfer metrics that are closely aligned
with our temporal proxy tasks, so that the main text focuses on the evaluation targets most
directly connected to the training objective. To make this selection transparent rather than
selective, Fig.~\ref{fig:app_mt_opd_extra_curves} reports the remaining four benchmarks from
the same eight-benchmark suite. Together with Fig.~\ref{fig:mt_opd_time_efficiency}, these
curves cover the full suite without duplicating panels between the main text and appendix.
The aggregate conclusion is consistent with the main text: averaged over all eight metrics,
EventLens obtains a +3.39 gain after 12.7 hours, compared with +2.87 for Mixed-GRPO at its nearest
wall-clock checkpoint (13.0 hours) and +3.38 for Mixed-GRPO after 36.7 hours.

The broad video QA benchmarks should be interpreted more cautiously than the targeted temporal
metrics. Video-MME and Video-Holmes show small, non-monotonic changes whose magnitudes are much
smaller than the gains on ETBench, TimeLens, and Vinoground. We therefore use the full-suite
curves across the main text and appendix as a transparency check rather than as evidence that
every benchmark improves
monotonically. The main empirical claim is wall-clock efficiency: under the same mixed
proxy-task pool, EventLens reaches a comparable average transfer level in roughly one third of the
training time required by direct Mixed-GRPO.

\section{Limitations}
\label{sec:limitations}

The coverage, granularity, and semantic accuracy of automatically constructed event chains determine the quality of the recovery supervision.
Our current construction audit uses model-assisted judgments, and the timestamp-perturbation experiment is a 100-step pilot.
Human validation and experiments that vary event semantics or hierarchy would extend the assessment of supervision quality.

The present evaluation covers Qwen3-VL and Qwen2.5-VL under the reported video-input settings.
Recovery tasks use event clips and local multi-event windows; extending them to longer narratives, denser temporal sampling, and higher-resolution motion is a natural next step.

The efficiency curves measure the MT-OPD student stage.
A complete compute account also includes task-specialized teacher training and teacher-distribution queries.
The consolidated student runs independently of its teachers at inference.

\end{document}